\documentclass{article}

\usepackage[english]{babel}

\usepackage[letterpaper,top=2cm,bottom=2cm,left=3cm,right=3cm,marginparwidth=1.75cm]{geometry}

\usepackage{amsmath}
\usepackage{booktabs}
\usepackage{epsfig}
\usepackage{multirow}
\usepackage{authblk}

\usepackage{tikz}
\usepackage[numbers]{natbib}

\usepackage{diagbox}
\usepackage{graphicx}
\usepackage{hyperref}
\hypersetup{colorlinks=true, allcolors=blue}

\title{deepFDEnet: A Novel Neural Network Architecture for Solving Fractional Differential Equations}
\author[$\dagger$]{Ali Nosrati Firoozsalari$^+$}
\author[$\dagger$]{Hassan Dana Mazraeh$^+$}
\author[$\dagger$]{Alireza Afzal Aghaei}
\author[$\dagger$]{Kourosh Parand\thanks{Corresponding author}}
\affil[$\dagger$]{Department of Computer and Data Sciences, Shahid Beheshti University, G.C., Tehran, Iran.}
\affil[ ]{a.nosratif@gmail.com, h\_danamazraeh@sbu.ac.ir, alirezaafzalaghaei@gmail.com, k\_parand@sbu.ac.ir}

\begin{document}
\maketitle
\def\thefootnote{+}\footnotetext{These are co-first authors and contributed equally to this work}\def\thefootnote{\arabic{footnote}}

\begin{abstract}
The primary goal of this research is to propose a novel architecture for a deep neural network that can solve fractional differential equations accurately. A Gaussian integration rule and a $L_1$ discretization technique are used in the proposed design. In each equation, a deep neural network is used to approximate the unknown function. Three forms of fractional differential equations have been examined to highlight the method's versatility: a fractional ordinary differential equation, a fractional order integrodifferential equation, and a fractional order partial differential equation. The results show that the proposed architecture solves different forms of fractional differential equations with excellent precision.

\end{abstract}

\section{Introduction}
Fractional equations are mathematical equations that include one or more fractional derivatives or fractional integral terms. These types of equations are encountered in many branches of science and engineering, such as physics, chemistry, and cognitive sciences \cite{G._2003,Mehdi}. One important application of fractional equations is in modeling physical systems that exhibit non-integer behavior, such as visco-elastic materials or fractional-order circuits. Fractional equations are challenging to solve because they often involve complex algebraic manipulations and require the use of specialized techniques such as partial fraction decomposition. In recent years, there has been a surge of interest in the study of fractional calculus, which extends classical calculus to include fractional derivatives and integrals and provides a powerful mathematical framework for the analysis and modeling of fractional equations. Since nowadays neural networks have been proven to be a powerful tool in solving differential equations, in this research paper we present a deep neural network (DNN) framework to solve fractional differential equations (FDE). DNNs are the proper choice to solve FDEs because traditional methods for FDEs can be computationally expensive, difficult to implement for complex problems, and might not scale well to high-dimensional systems. In addition, these days we encounter big data coming from different sensors that measure the boundary and initial conditions of a FDE under consideration. DNNs have also demonstrated a high capability for dealing with large amounts of data. As a result, examining DNNs ability to solve FDEs seems necessary, which is the primary focus of this paper.

Neural networks can learn the underlying dynamics of a system directly from the data. This is useful when one or more conditions (boundary or initial) of the equation under consideration are unknown or difficult to evaluate.
In recent years, a powerful DNN architecture called Physics-Informed Neural Networks (PINNs) has been proposed by M. Raissi et al. in \cite{Raisi}. PINNs are a class of neural networks that utilize physical laws, boundaries, and initial conditions in their architecture to improve the accuracy and reliability of their predictions. These networks use the partial differential equation (PDE) or ordinary differential equation (ODE) as the loss function. Furthermore, boundary and initial conditions are considered in the loss function as well. The total loss in these networks is as follows:
$$ Loss=Loss_{PDE}+Loss_{Boundries}+Loss_{Initial}.$$
The advantage of PINNs over traditional numerical methods is that they can learn from data and generalize to new forms of equations. This makes PINNs useful in situations where the governing equations are complex or where the initial and boundary conditions are uncertain or incomplete.

In this work, a fractional PINN has been proposed to overcome the difficulties arising in solving fractional differential equations.
Solving fractional PDEs using neural networks has been a pressing issue, and scientists have proposed different methods and techniques. In 2012, Almarashi proposed a neural network to approximate the solution of a two-sided fractional partial differential equation with RBF, and they presented their result in \cite{Marashi}.
Dai and Yu proposed an artificial neural network to solve space fractional differential equations with the Caputo definition by applying truncated series expansion terms. \cite{sym14030535}
In addition, Pang et al. proposed a neural network, used the Grünwald-Letnikov formula to discretize the fractional operator, and compared their convergence against the Finite Difference Method (FDM) \cite{doi:10.1137/18M1229845}.
Guoa et al. proposed a Monte Carlo Neural network for forward and inverse neural networks. They used a similar approach to fPINNs, however, their approach yields less overall computational
cost compared to fPINN \cite{GUO2022115523}.

The methods described above provide numerous approaches of solving fractional equations, each with its own set of advantages and disadvantages. Regardless, our suggested approach is accurate and versatile since it utilizes L1 discretization to discretize the fractional part of the equations and Gauss-Legendre integration to discretize the integral component. 

This research article has been divided into multiple key sections that will aid in a thorough comprehension of the proposed method. Section \ref{BG} provides background information necessary for understanding the proposed method. Section \ref{MG} will provide an overview of the methodology employed, and the results will be presented in Section \ref{NR}. Finally, in Section \ref{CN} we will present our concluding remarks and discuss further research.

\section{Background} \label{BG}

\subsection{Fractional Calculus}
In this paper, we primarily focus on fractional-order equations. There are different definitions for the fractional order differential equations. We will examine some of the most important definitions and then will introduce our proposed method, utilizing the Gauss-Legendre integration method and L1-discretization method.
Considering the Riemann–Liouville fractional integral with $f\in C[a,b]$ \: and \: $\alpha\in{R^{+}}$, we will have:
\begin{equation}
_a\mathcal{I}^\alpha_xf(x)=\frac{1}{\Gamma(\alpha)}\int_a^x(x-s)^{\alpha-1}f(s)ds,\end{equation}
In the above equation, the $\Gamma$ stands for the gamma function.
The Riemann–Liouville(RL) fractional integral can be formulated as follows\cite{Miller_Ross_1993}:

\begin{equation}\label{RL}
    _a\mathcal{D}^\alpha_xf(x)=\dfrac{1}{\Gamma(n-\alpha)}\dfrac{d^n}{dx^n}\int_a^x(x-s)^{n-\alpha-1}f(s)ds,
\end{equation}
Equation \ref{RL} shows the RL definition of fractional derivative. It can be observed that in this equation, the integral part is computed first, followed by the derivative; however, there is a more straightforward definition in which the derivative is calculated first, followed by the integral. This is known as the Caputo definition, and it goes as follows
\cite{10.1111/j.1365-246X.1967.tb02303.x}:
\begin{equation}
    _a D_t^\alpha f(x)=\dfrac{1}{\Gamma(n-\alpha)}\int_a^x(x-s)^{n-\alpha-1}f^{(n)}(s)ds,\quad n-1<\alpha\le n.
\end{equation}
According to the Caputo definition, for polynomials, we have:
\begin{equation}
\partial_0^\alpha x^n=\begin{cases}0&\text{n}\in\mathrm{N}_0,\text{n}<\lceil\alpha\rceil,\\ \frac{\Gamma(\text{n}+1)}{\Gamma(n+1-\alpha)}x^{n-\alpha}&\text{n}\in\mathrm{N},\text{n}\geq\lceil\alpha\rceil.\end{cases}
\end{equation}
The Caputo definition with $0<\alpha\leq1$ initiating at zero can be formulated as:
\begin{equation}
    \begin{aligned}\partial_t^\alpha u(x,t)=\frac{1}{\Gamma(1-\alpha)}\int_0^t\frac{1}{(t-s)^\alpha}\frac{\partial u(x,s)}{\partial s}ds.\end{aligned}
\end{equation}

\subsection{L1-discretization}
Because solving the Caputo fractional equation is computationally difficult, numerous scientists have proposed alternative ways for determining the fractional derivative. The L-1 and L1-2 discretization methods are two that are used to effectively calculate the fractional derivative.
Using L-1 discretization to approximate the Caputo derivative where $t= t_{n}+1$, will result in the following expression:
\begin{equation}
    \partial_t^\alpha u\left(x,t_{n+1}\right)=\mu(u^{n+1}-(1-b_1)u^n-\sum_{j=1}^{n-1}(b_j-b_{j+1})u^{n-j}-b_ku^0)+r^{n+1}_{\Delta t},\quad n\geq1,
\end{equation}
Where $\alpha$ is the non-integer order, $\mu=\frac{1}{\Delta t^\alpha\Gamma(2-\alpha)}$ and $b_j=(j+1)^{1-\alpha}-j^{1-\alpha}$. This is known as the L-1 discretization, which is used in our proposed deep network.

\section{Methodology} \label{MG}
This section will cover the Gauss-Legendre integration first, followed by an in-depth examination of the proposed method.

\subsection{Gauss-Legendre integration}
There are different methods to approximate the definite integral function numerically, some of which are the midpoint rule, trapezoidal rule\cite{Ossendrijver_2016}, Simpson's rule\cite{Atkinson_1989}, and the Gaussian quadrature\cite{gauss_2011}. In this paper, we have utilized the Gaussian Quadrature and we will briefly explain this method in the following paragraph.
The Gaussian quadrature method used in this paper is based on the formulation that was developed by Carl Gustav Jacobi in 1826.
The default domain for integration in the Gaussian quadrature rule is usually taken as [-1,1] and it is stated as:
\begin{equation}
    {\displaystyle \int _{-1}^{1}w(x)\,dx\approx \sum _{i=1}^{n}w_{i}f(x_{i}),}
\end{equation}

if the function $f(x)$ can be approximated by a polynomial of degree $2n - 1$, then the integral calculated by the above rule is an accurate approximation. It should also be taken into consideration that the points for assessment in the Gaussian quadrature are chosen optimally and are not equally spaced. These points are the roots of Legendre polynomials of degree $n$, which is the number of points used to approximate the solution.

Since the integrals considered in this paper are not all in $[-1,1]$, we make use of the the following transformation which leverages the change of variables to calculate the integral in any arbitrary interval.
\begin{equation}
    t=\frac{2x-a-b}{b-a}\Longleftrightarrow x=\frac{1}{2}[(b-a)t+a+b],
\end{equation}
this will result in the following formula for the Gaussian quadrature:

\begin{equation}
    \int_a^b w(x)dx=\int_{-1}^1f\left(\frac{(b-a)t+(b+a)}{2}\right)\frac{(b-a)}{2}dt.
\end{equation}

\subsection{Proposed Method}
We have already covered the necessary material to understand the proposed method. We employ a sequential Neural Network and to solve the equations, we use tanh(.) as activation functions, moreover, we assume that our network will calculate the result as $U(.)$, then to find the result of our equation, we utilize automatic differentiation to find the derivatives with respect to each variable. We then use L1-discretization for the fractional derivative and Gauss-Legendre to calculate the integral part.
Consider the equation of the form:

\begin{figure}
    \centering
    \includegraphics[width=1\textwidth]{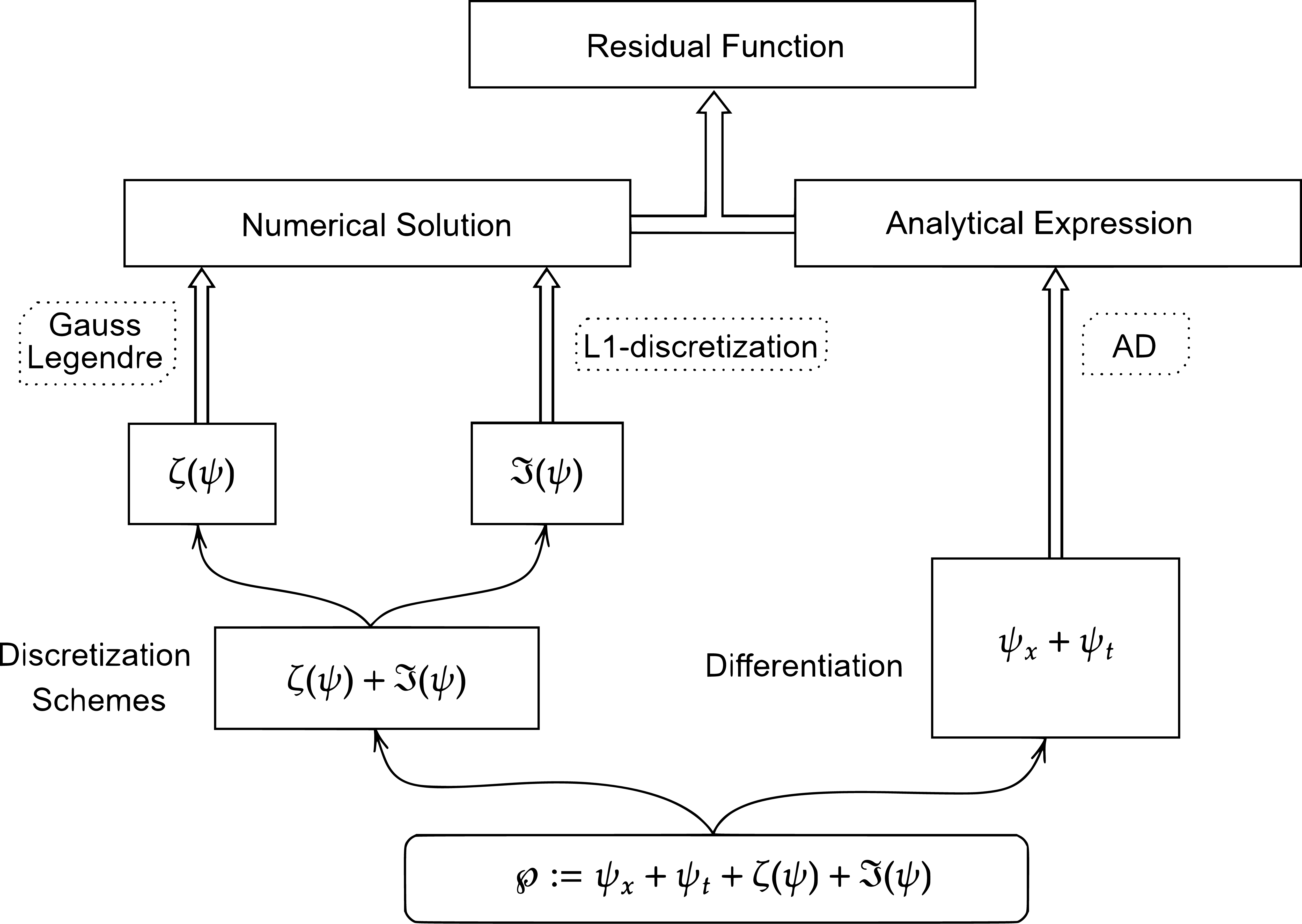}
    \caption{A diagram illustrating the proposed method}
    \label{fig:enter-label}
\end{figure}
\begin{equation}
    \wp:=\psi+\psi_t+\zeta(\psi)+\Im(\psi),
\end{equation}
where:
\begin{equation}
    \zeta := \sum _{i=1}^{n}w_{i}f(\psi_{i}),\quad n\geq1,
\end{equation}
and:
\begin{equation}
    \Im := \mu(\psi^{n+1}-(1-b_1)\psi^n-\sum_{j=1}^{n-1}(b_j-b_{j+1})\psi^{n-j}-b_k\psi^0)+r^{n+1}_{\Delta t},\quad n\geq1,
\end{equation}
where $\psi(x,t)$ is calculated by the neural network, and the derivative with respect to each variable is calculated using automatic differentiation, however, the integral is calculated using the Gauss-Legendre method as is shown by $\zeta$ and the fractional derivative is estimated using L1-discretization and it is shown by $\Im$. In each iteration, a random point is generated and the value is calculated using the neural network, then the random value is chosen as the maximum value in both the integration scheme and to estimate the fractional derivative. The equation is considered as part of the loss function and the parameters are learned using the squared error of the values in each equation and initial and boundary conditions. The shared parameters will be learned by minimizing $SE$ as follows:
\begin{equation}
    SE = SE_\Im + SE_i + SE_b,
\end{equation}

where:

\begin{equation}
    SE_\Im = |\psi(t_{\Im}^{0},x_{\Im}^{0})|^{2},
\end{equation}
\begin{equation}
    SE_i = |\psi(t_{0}^{i},x_{0}^{i})-\psi_{0}^{i}|^{2},
\end{equation}

and
\begin{equation}
    SE_b = |\psi(t_{b}^{i},x_{b}^{i})-\psi_{b}^{i}|^{2}.
\end{equation}
Figure \ref{fig:enter-label} illustrates the diagram of the proposed method.

\section{Numerical Results} \label{NR}
In this section, some examples are investigated and solved using the given methods to demonstrate the efficacy of the proposed method. To demonstrate the generality of the proposed method, we have selected some different examples from different classes of fractional order differential equations including a fractional order differential equations, a fractional order integrodifferential equation, and a fractional order partial differential equation.

\subsection{Example 1}
\begin{figure}
\begin{minipage}{.5\textwidth}
    \centering
    \includegraphics[width=1\textwidth]{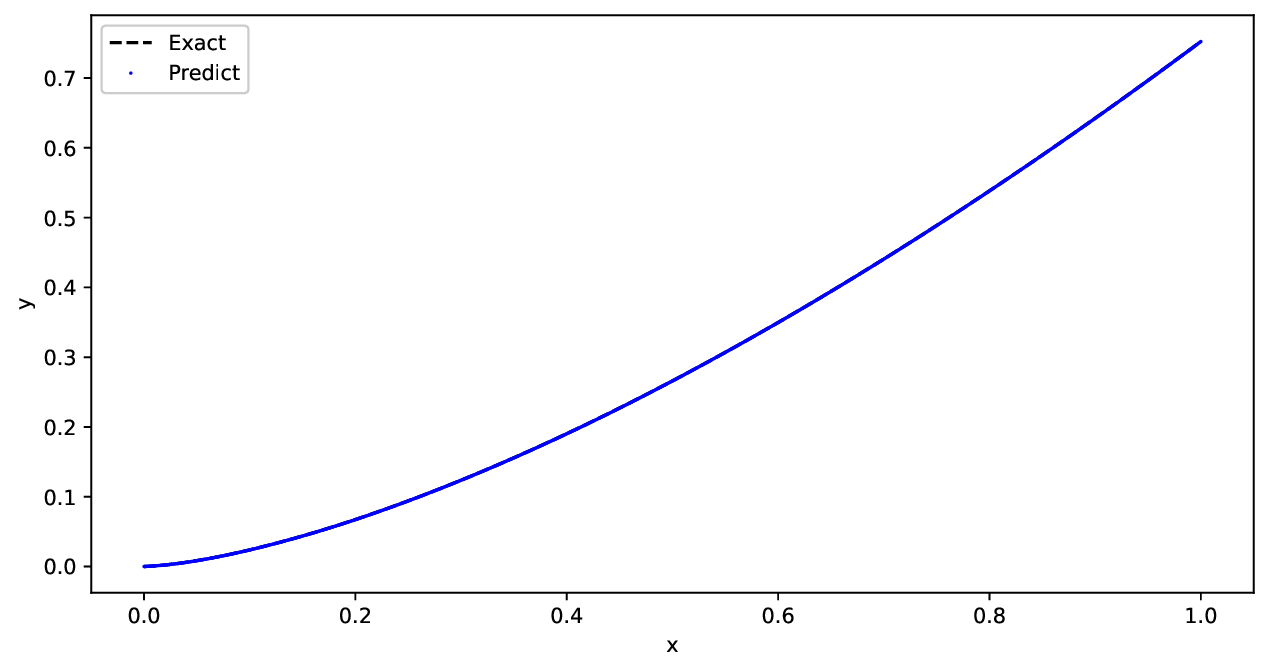}
    \caption{Predicted solution of example 1.}
    \label{fig1-1}
\end{minipage}
\begin{minipage}{.5\textwidth}
    \centering
    \includegraphics[width=1\textwidth]{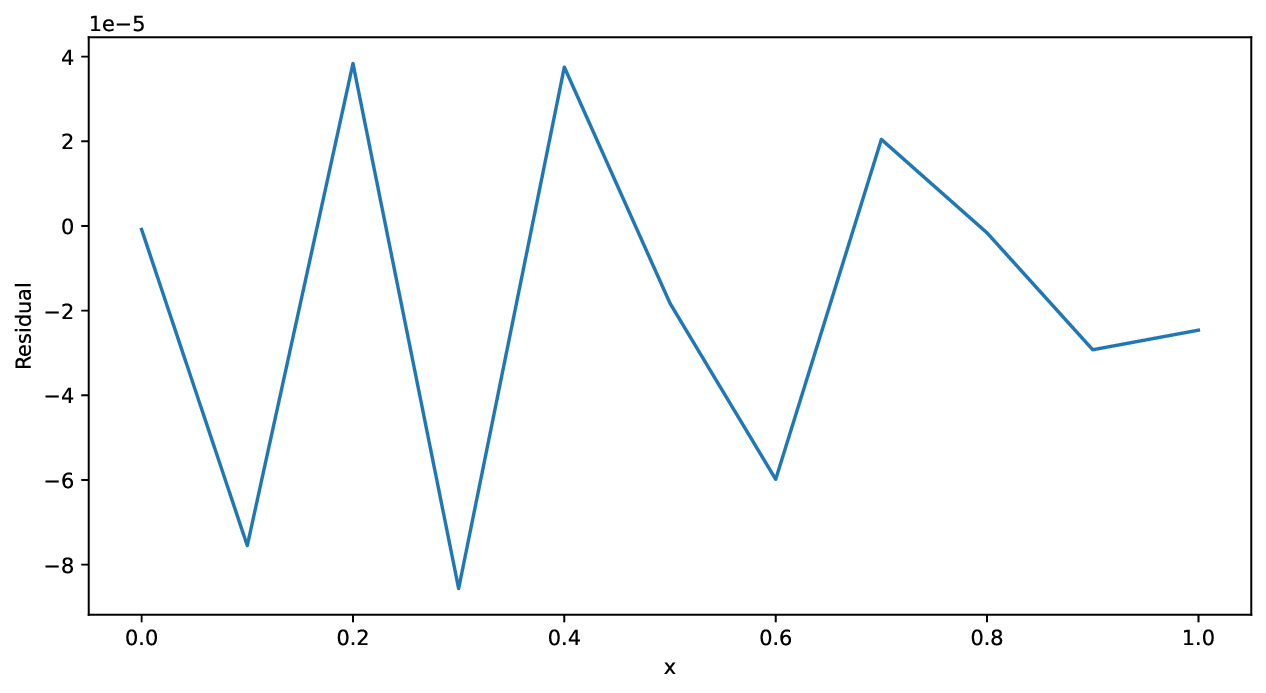}
    \caption{Residual graph for example 1.}
    \label{fig1-2}
\end{minipage}
\end{figure}

First, we consider a fractional ordinary differential equation as follows\cite{RAHIMKHANI20168087}:

\begin{equation}
    D^\nu \psi(x)+\psi^2(x)=x+\left(\frac{x^{\nu+1}}{\Gamma(\nu+2)}\right)^2,\quad0<\nu\le1,\quad0\le x\le1.
\end{equation}
The initial condition for this equation is $\psi(0)=0$ and the exact solution is $\psi(x) = \frac{1}{\Gamma(v+2)} x^{\nu+1}.$
The residual will consist of the initial condition and the equation itself:
\begin{equation}
    SE = SE_\Im + SE_i,
\end{equation}
where
\begin{equation}
    SE_\wp=\Im(\psi)+\psi^2-(x+\left(\frac{x^{\nu+1}}{\Gamma(\nu+2)}\right)^2),
\end{equation}
and
\begin{equation}
    SE_i = |\psi({0})-0|.
\end{equation}
The results are presented in table \ref{ex1-1}, where the mean absolute error is calculated for different values for $\alpha$ with $n=1000$ discretization points and $1000$ epochs. Figure \ref{fig1-1} depicts the exact and predicted solution to the presented problem and figure \ref{fig1-2} shows the residual. 


\begin{table}[]
\centering
\begin{tabular}{@{}cccccc@{}}
\toprule
\diagbox{x}{$\alpha$}   & 0.1       & 0.3       & 0.5       & 0.7       & 0.9       \\ \midrule
0   & 3.23E-05  & -7.14E-06 & -6.25E-06 & 3.43E-05  & -6.31E-05 \\
0.1 & -1.45E-05 & 7.75E-05  & -1.86E-05 & 4.12E-05  & -1.11E-04 \\
0.2 & 2.51E-04  & -5.37E-05 & -1.42E-05 & -3.04E-05 & -7.69E-05 \\
0.3 & -8.31E-05 & -6.13E-05 & -4.44E-05 & -1.49E-05 & -1.42E-04 \\
0.4 & -1.25E-04 & -2.33E-05 & -1.92E-05 & 1.21E-05  & -2.15E-04 \\
0.5 & 1.56E-05  & 2.59E-05  & 8.61E-06  & -1.89E-06 & -2.37E-04 \\
0.6 & 9.12E-05  & 2.40E-06  & -2.46E-05 & -2.39E-05 & -2.12E-04 \\
0.7 & 4.85E-05  & -3.96E-05 & -2.54E-05 & -2.42E-05 & -1.83E-04 \\
0.8 & -2.86E-05 & -2.48E-05 & 9.58E-06  & -1.28E-05 & -1.91E-04 \\
0.9 & -3.38E-05 & 1.59E-05  & -2.11E-05 & -1.31E-05 & -2.31E-04 \\
1   & 1.18E-04  & -3.73E-05 & 2.95E-05  & -1.11E-05 & -2.11E-04 \\ \bottomrule
\end{tabular}
\caption{Mean absolute error for different values of $\alpha$ and $1000$ epochs for example 1.}
\label{ex1-1}
\end{table}




\subsection{Example 2}
\begin{figure}
\begin{minipage}{.5\textwidth}
    \centering
    \includegraphics[width=1\textwidth]{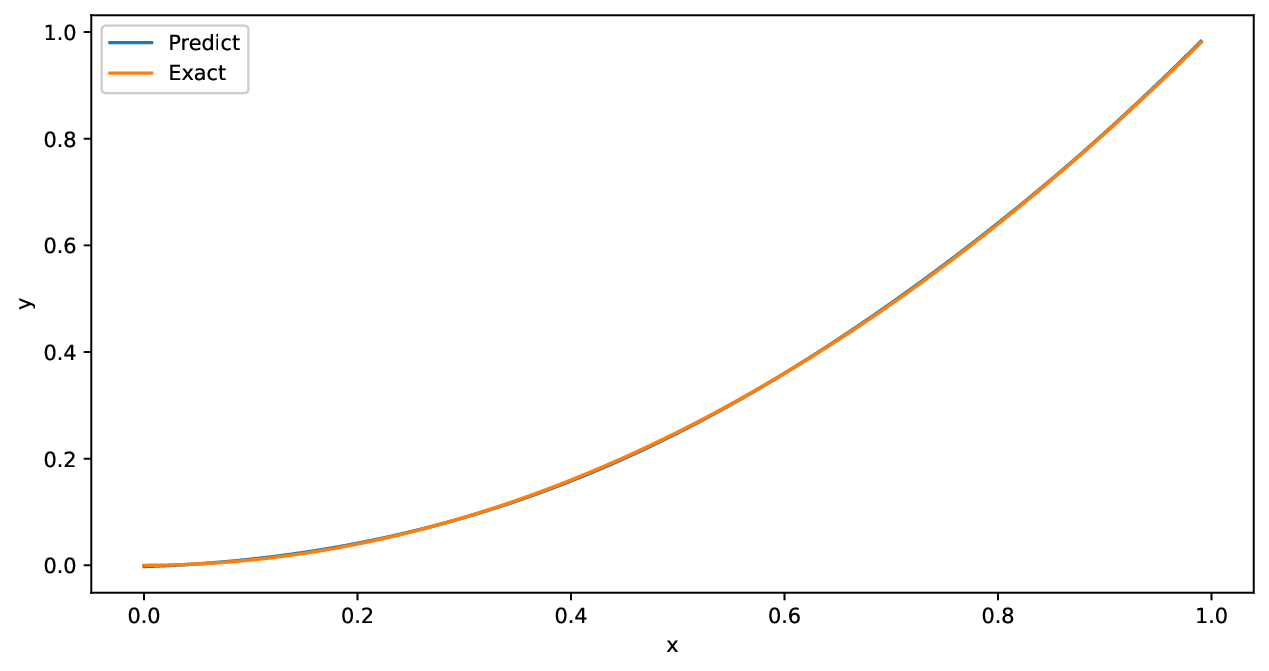}
    \caption{Predicted solution of example 2.}
    \label{fig2-1}
\end{minipage}
\begin{minipage}{.5\textwidth}
    \centering
    \includegraphics[width=1\textwidth]{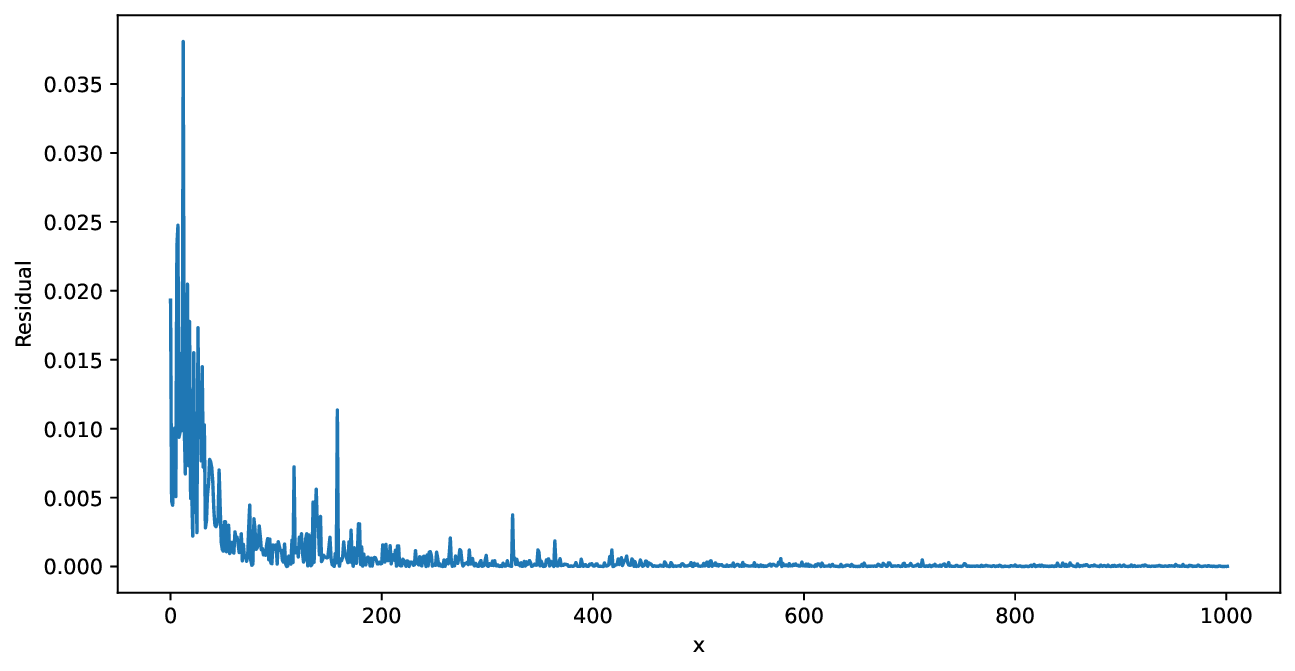}
    \caption{Residual graph for example 2.}
    \label{fig2-2}
\end{minipage}
\end{figure}

now we consider a fractional integral equation\cite{TOMA2023469}:

\begin{equation}
    \left\{\begin{array}{l}{{D^{0.5}\psi\left(x\right)=\psi\left(x\right)+\frac{8}{\sin\left(0.5\right)}x^{1.5}-x^{2}-\frac{1}{3}x^{3}+\displaystyle\int_{0}^{x}\psi\left(t\right)d t,\quad0<x<1,}}\\ {{\psi\left(0\right)=0,}}\end{array}\right.
\end{equation}
The solution to this equation is $\psi(x)=x^{2}.$
The residual will consist of the initial condition and the equation itself:
\begin{equation}
    SE = SE_\Im + SE_i,
\end{equation}
where
\begin{equation}\label{ex.2}
    SE_\wp=\Im(\psi)-(\psi(x)+\frac{8}{\sin\left(0.5\right)}x^{1.5}-x^{2}-\frac{1}{3}x^{3}+\zeta(\psi)),
\end{equation}
and
\begin{equation}
    SE_i = |\psi({0})-0|,
\end{equation}
and the results are as presented in table \ref{tab2-1} with varying numbers of discretization points. figure \ref{fig2-1} depicts the predicted and exact solutions, whereas figure \ref{fig2-2} shows the residual graph for $1000$ epochs. As shown in equation \ref{ex.2}, this equation consists of a fractional part and an integral part. The fractional component is calculated with various numbers of discretization points as shown in table \ref{tab2-1}, however, we have utilized $400$ discretization points in the Gauss-Legendre method to estimate the integral part.



\begin{table}[]
\centering
\begin{tabular}{@{}cccccc@{}}
\toprule
\diagbox{x}{k}   & 100      & 250      & 500      & 750      & 1000     \\ \midrule
0   & 2.22E-03 & 1.19E-03 & 1.30E-03 & 5.94E-04 & 2.26E-03 \\
0.1 & 1.01E-02 & 6.55E-03 & 1.64E-03 & 2.36E-03 & 1.44E-03 \\
0.2 & 1.02E-02 & 7.02E-03 & 2.62E-03 & 2.47E-03 & 1.42E-03 \\
0.3 & 6.48E-03 & 4.97E-03 & 2.40E-03 & 1.46E-03 & 2.94E-04 \\
0.4 & 2.45E-03 & 2.46E-03 & 2.17E-03 & 6.45E-04 & 1.75E-03 \\
0.5 & 3.29E-04 & 8.79E-04 & 3.02E-03 & 6.67E-04 & 1.58E-03 \\
0.6 & 2.24E-04 & 5.26E-04 & 4.87E-03 & 1.41E-03 & 1.72E-04 \\
0.7 & 4.38E-04 & 6.07E-04 & 5.92E-03 & 2.22E-03 & 1.87E-03 \\
0.8 & 5.05E-04 & 1.80E-04 & 4.47E-03 & 2.57E-03 & 1.76E-03 \\
0.9 & 2.73E-03 & 8.18E-04 & 1.89E-03 & 2.72E-03 & 1.09E-03 \\
1   & 9.65E-03 & 8.60E-03 & 7.86E-04 & 1.90E-03 & 2.16E-03 \\ \bottomrule
\end{tabular}
\caption{Mean absolute error for example 2 with different discretization points and $1000$ epochs.}
\label{tab2-1}
\end{table}

\subsection{Example 3}
Finally, we consider the initial-boundary problem of fractional partial differential equation\cite{SAADATMANDI20124125}:
\begin{equation}
    \frac{\partial^\alpha \psi(x,t)}{\partial t^\alpha}+x\frac{\partial \psi(x,t)}{\partial x}+\frac{\partial^2\psi(x,t)}{\partial x^2}=2t^\alpha+2x^2+2,\quad0<x<1,0<t<1,
\end{equation}
where $\alpha$ is between $0$ and $1$, the initial condition is $\psi(x,0)=x^2$ and the boundary conditions are:

$\psi(0,t)=2\frac{\Gamma(\alpha+1)}{\Gamma(2\alpha+1)}t^{2\alpha},$ and $\psi(1,t)=1+2\frac{\Gamma(\alpha+1)}{\Gamma(2\alpha+1)}t^{2\alpha}.$
\\The residual will consist of the initial condition and the equation itself as follows:
\begin{equation}
    SE = SE_\wp + SE_i + SE_b,
\end{equation}
where
\begin{equation}
    SE\wp=\Im(\psi)+\psi_x+\psi_{xx}-(2t^\alpha+2x^2+2),
\end{equation}
and
\begin{equation}
    SE_i = |\psi({x,0})-x^2|,
\end{equation}

\begin{equation}
    SE_b = |\psi(0,x^{i})-2\frac{\Gamma(\alpha+1)}{\Gamma(2\alpha+1)}t^{2\alpha}|^{2} + |\psi(1,x^{i})-(1+2\frac{\Gamma(\alpha+1)}{\Gamma(2\alpha+1)}t^{2\alpha})|^{2},
\end{equation}
\begin{table}[]
\centering
\begin{tabular}{@{}cccc@{}}
\toprule
\multirow{2}{*}{x} & \multicolumn{2}{c}{Wavelet Method}\cite{schumer2009fractional} & \multirow{2}{*}{Presented Method} \\ \cmidrule(lr){2-3}
                   & m=32             & m=64            &                                   \\ \cmidrule(r){1-1} \cmidrule(l){4-4} 
0.1                & 6.09E-03         & 1.21E-03        & 8.43E-03                          \\
0.2                & 4.84E-03         & 1.25E-03        & 8.45E-03                          \\
0.3                & 2.75E-02         & 1.86E-03        & 7.64E-03                          \\
0.4                & 1.93E-02         & 7.41E-03        & 6.19E-03                          \\
0.5                & 1.00E-06         & 1.00E-06        & 4.45E-03                          \\
0.6                & 4.35E-02         & 7.46E-03        & 2.73E-03                          \\
0.7                & 1.73E-02         & 1.72E-03        & 1.23E-03                          \\
0.8                & 7.75E-02         & 4.99E-03        & 8.98E-05                          \\
0.9                & 4.44E-02         & 1.67E-02        & 5.91E-04                          \\ \bottomrule
\end{tabular}
\caption{Comparison of mean absolute error between Wavelet Method and our method for $\alpha$=0.5 and t=0.5 in example 3.}
\label{tab3-1}
\end{table}

\begin{figure}
    \centering
    \includegraphics[width=1\textwidth]{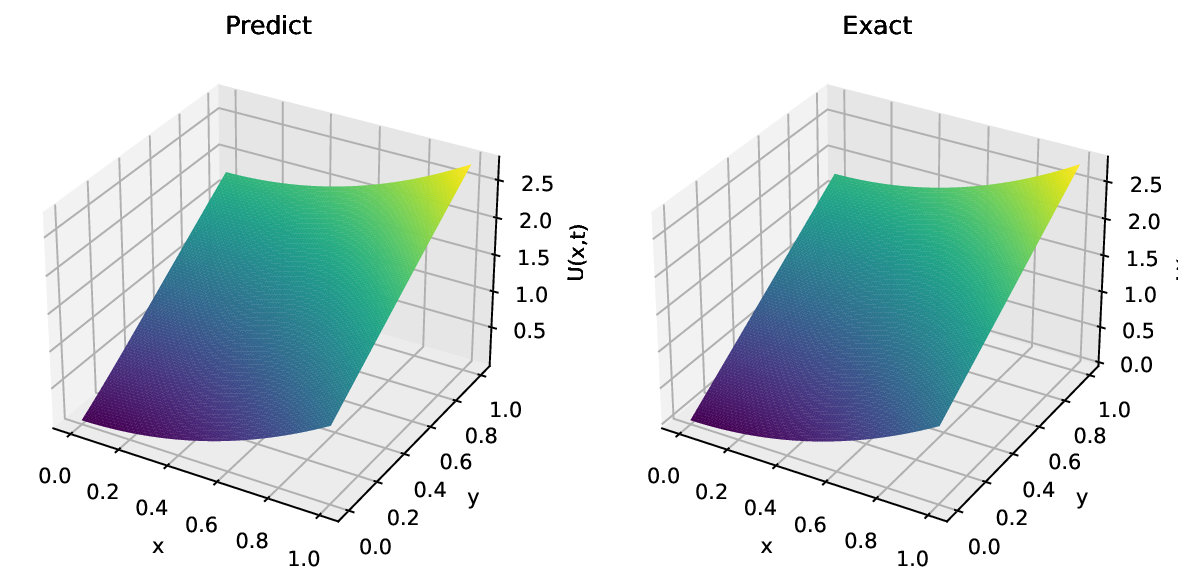}
    \caption{Predicted solution and the exact solution for example 3.}
    \label{fig3}
\end{figure}

\begin{table}[]
\centering
\begin{tabular}{@{}cccccc@{}}
\toprule
\diagbox{x}{t}   & 0.1      & 0.3      & 0.5      & 0.7      & 0.9      \\ \midrule
0   & 8.07E-03 & 1.65E-02 & 7.71E-03 & 3.81E-03 & 1.13E-02 \\
0.1 & 6.34E-03 & 1.59E-02 & 8.43E-03 & 2.49E-03 & 1.07E-02 \\
0.2 & 4.67E-03 & 1.44E-02 & 8.45E-03 & 1.12E-03 & 8.95E-03 \\
0.3 & 2.91E-03 & 1.21E-02 & 7.64E-03 & 7.14E-05 & 6.59E-03 \\
0.4 & 1.11E-03 & 9.19E-03 & 6.19E-03 & 6.34E-04 & 3.93E-03 \\
0.5 & 6.15E-04 & 6.10E-03 & 4.45E-03 & 1.14E-03 & 1.09E-03 \\
0.6 & 2.09E-03 & 3.18E-03 & 2.73E-03 & 1.58E-03 & 1.85E-03 \\
0.7 & 3.15E-03 & 6.91E-04 & 1.23E-03 & 2.07E-03 & 4.88E-03 \\
0.8 & 3.67E-03 & 1.19E-03 & 8.98E-05 & 2.70E-03 & 8.01E-03 \\
0.9 & 3.51E-03 & 2.36E-03 & 5.91E-04 & 3.60E-03 & 1.14E-02 \\
1   & 2.56E-03 & 2.73E-03 & 6.55E-04 & 4.98E-03 & 1.52E-02 \\ \bottomrule
\end{tabular}
\caption{Mean absolute error for example 3 with different values for $x$ and $t$ and $1000$ epochs.}
\label{tab3-2}
\end{table}
The results are presented in table \ref{tab3-2}  and figure \ref{fig3} shows the approximate solution obtained by the current method using $n=100$ discretization points along with the exact solution. the Comparison between the obtained solutions and the Wavelet method is also provided in table \ref{tab3-1} which demonstrates that the current method can reliably evaluate the solution to the presented problem.
\section{Conclusion} \label{CN}
Solving the fractional differential equations using neural networks is a challenging task, especially considering that many fractional definitions have a singular kernel and cannot be directly computed. In this paper, we proposed a novel framework to solve the fractional differential equations using L1-discretization and the Gauss-Legendre discretization to discretize the integral component. The method can be used in several different fields including biological systems, medical imaging, stock prices, and control systems \cite{2006449}.
To demonstrate the effectiveness of the model, we considered several fractional equations including an ODE, a PDE, and an integro-differential equation. Solving fractional equations using neural networks is a relatively new research area and, while the previous works have contributed to this field, a solid framework for obtaining effective solutions to these equations remains lacking. By utilizing the aforementioned methodologies, we have developed a new and reliable framework to calculate the solutions to these equations. While our work mostly focuses on fractional equations, the usage of the Gauss-Legendre method can be expanded to the broader integral equations as well; furthermore, other discretization methods including L1-2 which is another discretization method for estimating the value of Caputo-type fractional equations, Gauss-Lobatto rule and adaptive quadrature for estimating the solution of integrals can also be considered and investigated. Our findings demonstrate that different discretization methods can be efficiently incorporated into a neural network; nevertheless, the effectiveness and usability of the method can be affected by parameters such as the depth of the neural network, activation functions, and the overall structure of the network. Furthermore, the effectiveness of different discretization methods in solving various fractional definitions is yet another subject that can be investigated in future work, nonetheless, the choice of the specific discretization method may depend on the specific characteristics of the equation and the structure of the neural network, notably the order of the fractional equation, the convergence order of the method and the computational complexity of the method which could in some cases severely hinder the speed and performance of our model. Finally, our  findings suggest that the proposed method and in general, discretization methods, are very valuable and could serve as a strong foundation for further research in this area.

\bibliographystyle{unsrt}
\bibliography{sample}

\end{document}